\documentclass[10pt,twocolumn,letterpaper]{article}

\usepackage{cvpr}
\usepackage{times}
\usepackage{epsfig}
\usepackage{graphicx}
\usepackage{amsmath}
\usepackage{amssymb}
\usepackage{xspace}

\usepackage{multirow}
\usepackage{tabularx}
\usepackage{colortbl}
\usepackage{booktabs}
\usepackage[table]{xcolor}
\usepackage{caption}

\usepackage{enumitem}

\usepackage[numbers,sort]{natbib}

\usepackage{xspace}
\usepackage{bbold}
\usepackage{xcolor}

\usepackage{eqparbox}
\newcommand{\mathbox}[3][\mathop]{%
  #1{\eqmakebox[#2]{$\displaystyle#3$}}%
}

\newcommand{\vct}[1]{\ensuremath{\boldsymbol{#1}}}

\newcommand{\set}[1]{\ensuremath{\mathcal{#1}}}

\usepackage{pifont}
\newcommand{\cmark}{\ding{51}}%
\newcommand{\xmark}{\ding{55}}%

\renewcommand{\paragraph}[1]{
        \vspace{1pt}
	\noindent\textbf{#1}}

\PassOptionsToPackage{hyphens}{url}\usepackage[pagebackref=true,breaklinks=true,letterpaper=true,colorlinks,bookmarks=false]{hyperref}

\cvprfinalcopy 

\def\cvprPaperID{5452} 

\ifcvprfinal\fi
\begin{document}

\title{Learning Multi-Object Tracking and Segmentation from Automatic Annotations}

\author{Lorenzo Porzi$^\dagger$, Markus Hofinger$^\ddagger$, Idoia Ruiz$^\ast$, Joan Serrat$^\ast$, Samuel Rota Bul\`o$^\dagger$, Peter Kontschieder$^\dagger$\\
Mapillary Research$^\dagger$, Graz University of Technology$^\ddagger$, Computer Vision Center, UAB$^\ast$\\
{\tt\small research@mapillary.com$^\dagger$, markus.hofinger@icg.tugraz.at$^\ddagger$, \{iruiz,joans\}@cvc.uab.es$^\ast$}
}

\newcommand{\KR}{KITTI Raw\xspace}
\newcommand{\KM}{KITTI MOTS\xspace}
\newcommand{\KS}{KITTI Synth\xspace}
\newcommand{\motsnet}{MOTSNet\xspace}

\def\dist{-1em}

\twocolumn[{%
\renewcommand\twocolumn[1][]{#1}%
\maketitle

\begin{center}
  \includegraphics[width=0.24\textwidth]{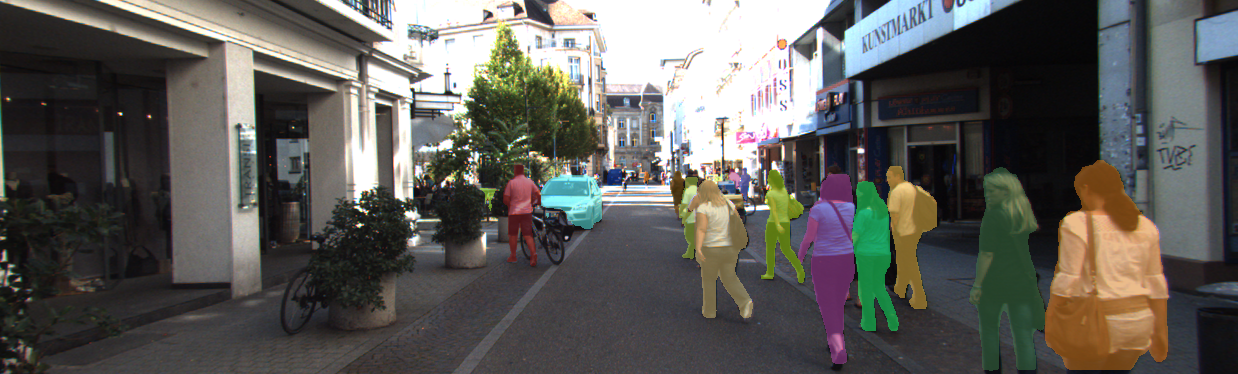}
  \includegraphics[width=0.24\textwidth]{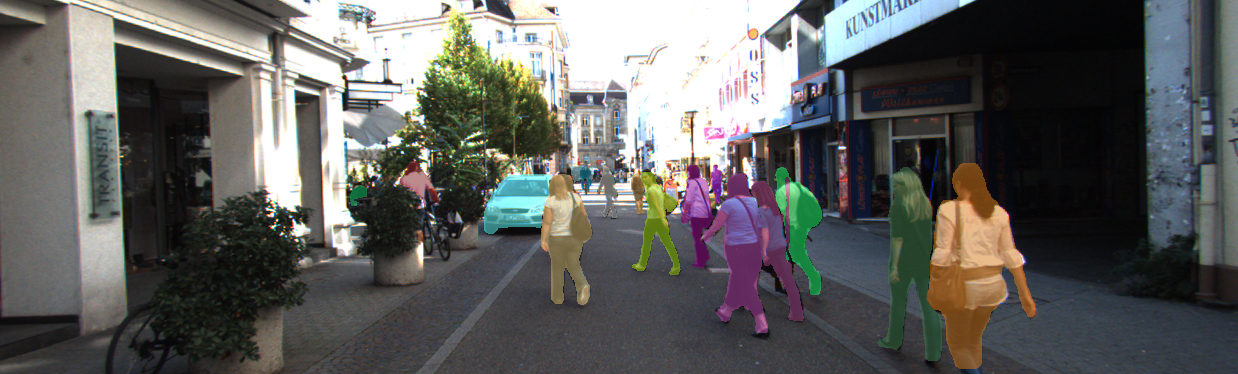}
  \includegraphics[width=0.24\textwidth]{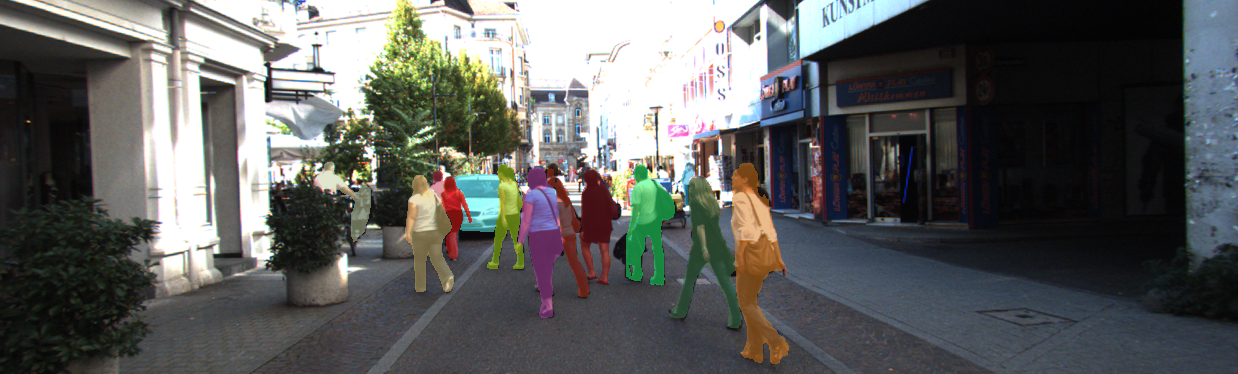}
  \includegraphics[width=0.24\textwidth]{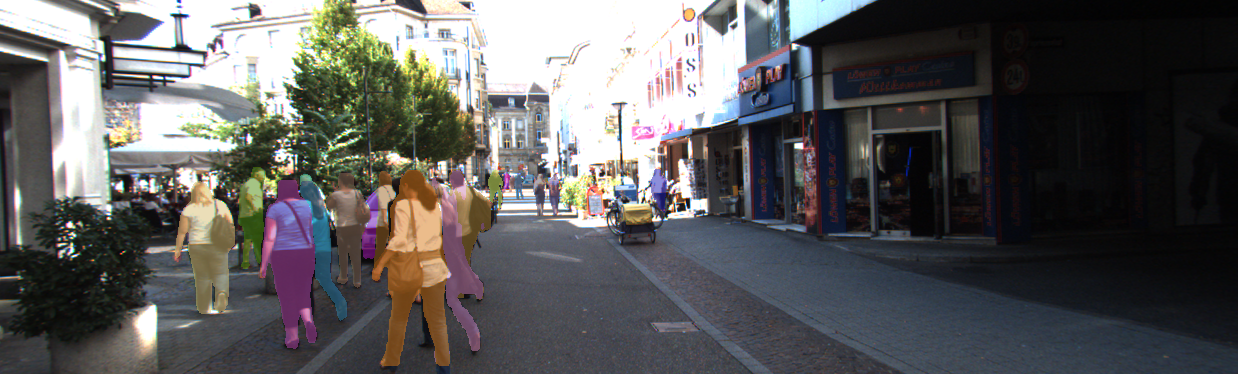}\\
  \includegraphics[width=0.24\textwidth]{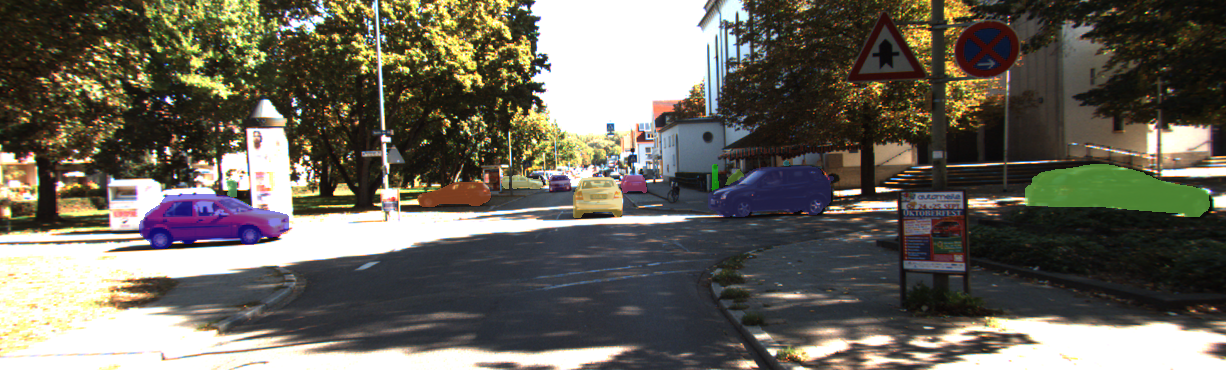}
  \includegraphics[width=0.24\textwidth]{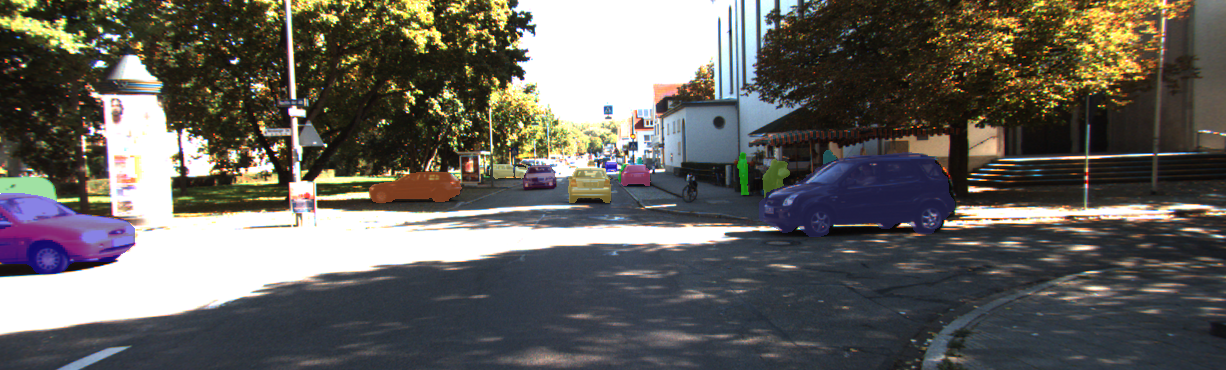}
  \includegraphics[width=0.24\textwidth]{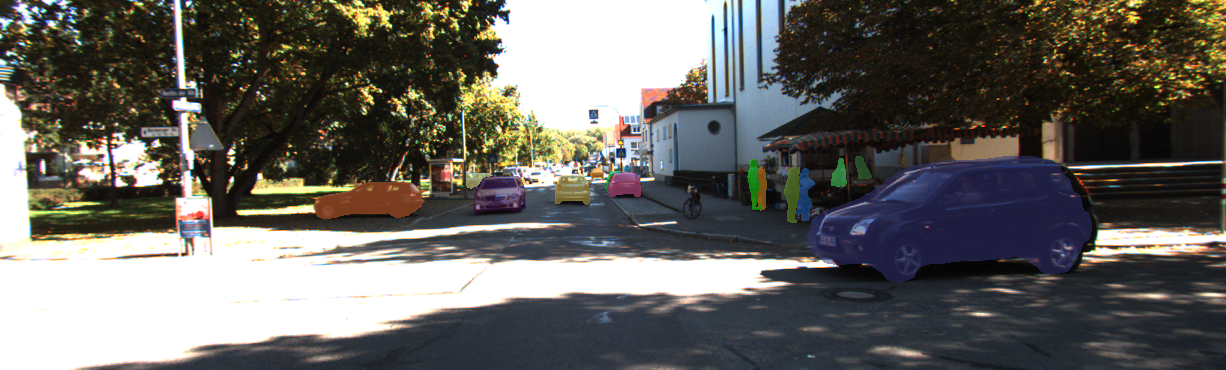}
  \includegraphics[width=0.24\textwidth]{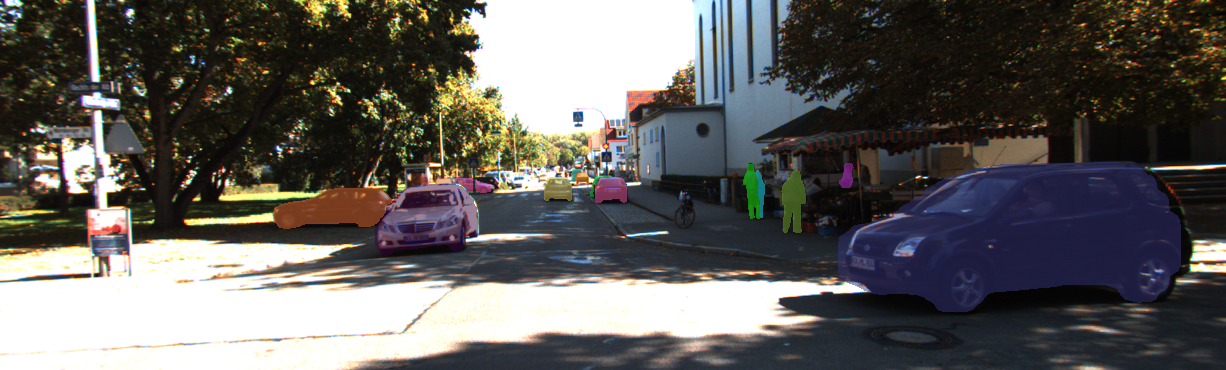}
  \captionof{figure}{KITTI sub-sequences with automatically generated MOTS annotations as color-coded instance masks (left to right).}
  \label{fig:sequence}
\end{center}
}]

\maketitle

\definecolor{mapillarygreen}{RGB}{5,203,99}

\begin{abstract}
In this work we contribute a novel pipeline to automatically generate training data, and to improve over state-of-the-art multi-object tracking and segmentation (MOTS) methods. Our proposed track mining algorithm turns raw street-level videos into high-fidelity MOTS training data, is scalable and overcomes the need of expensive and time-consuming manual annotation approaches.
We leverage state-of-the-art instance segmentation results in combination with optical flow predictions, also trained on automatically harvested training data. Our second major contribution is \motsnet -- a deep learning, tracking-by-detection architecture for MOTS -- deploying a novel mask-pooling layer for improved object association over time.
Training \motsnet with our automatically extracted data leads to significantly improved sMOTSA scores on the novel \KM dataset (+1.9\%/+7.5\% on cars/pedestrians), and \motsnet improves by +4.1\% over previously best methods on the MOTSChallenge dataset. Our most impressive finding is that we can improve over previous best-performing works, even in complete absence of manually annotated MOTS training data.
\end{abstract}

\section{Introduction and Motivation}
\label{sec:intro}
We focus on the challenging task of multi-object tracking and segmentation (MOTS)~\cite{Voigtlaender_2019_CVPR}, which was recently introduced as an extension of bounding-box based multi-object tracking.
Joining instance segmentation with tracking was shown to noticeably improve tracking performance, as unlike bounding boxes, instance segmentation masks do not suffer from overlapping issues but provide fine-grained, pixel-level information about objects to be tracked. 

While this finding is encouraging, it comes with the downside of requiring pixel-level (typically polygon-based) annotations, which are known to be time-consuming in generation and thus expensive to obtain.
The works in~\cite{Cordts2016,MVD2017} report annotation times of $\approx$90 minutes per image, to manually produce high-quality panoptic segmentation masks.
Analogously, it is highly demanding to produce datasets for the MOTS task where instance segmentation masks need to also contain tracking information across frames. 

To avoid generating MOTS labels completely from scratch and in a purely manual way, \cite{Voigtlaender_2019_CVPR} has followed the semi-automatic annotation procedure from~\cite{Caelles_2017_CVPR}, and extended the existing multi-object tracking datasets \textit{KITTI tracking}~\cite{Geiger2012CVPR} (\ie 21 sequences from the raw KITTI dataset) and MOTChallenge~\cite{MilanL0RS16} (4/7 sequences).
These datasets already provide bounding box-based tracks for cars and pedestrians.
Instance segmentation masks were generated as follows.
First, two segmentation masks per object were generated by human annotators (which also had to chose the objects based on diversity in the first place).
Then, DeepLabV3+~\cite{Chen2018ECCV}, \ie a state-of-the-art semantic segmentation network, was trained on the initially generated masks in a way to overfit these objects in a sequence-specific way, thus yielding reasonable segmentation masks for the remaining objects in each track.
The resulting segmentation masks then underwent another manual correction step to fix remaining errors.
Finally, these automatic and manual steps were iterated until convergence.

Including an automated segmentation algorithm is clearly speeding up the MOTS dataset generation process, but still has significant shortcomings.
Most importantly, the approach in~\cite{Voigtlaender_2019_CVPR} still depends on the availability of bounding-box based, multi-object tracking information as provided by datasets like~\cite{Geiger2012CVPR,MilanL0RS16}.
Also, their remaining human annotation effort for generating initial instance masks was considerable, \ie~\cite{Voigtlaender_2019_CVPR} reports that $\approx$8k masks (or $12.3\%$ of all masks in \KM) had been manually labeled before fine-tuning a (modified) DeepLabV3+ segmentation network, pre-trained on COCO~\cite{LinMSCOCO2014} and Mapillary Vistas~\cite{MVD2017}.
Finally, such an approach cannot be expected to generalize well for generating orders of magnitude more training data.

In this paper we introduce a novel approach for automatically generating high-quality training data (see Fig.~\ref{fig:sequence} for an example) from generic street-level videos for the task of joint multi-object tracking and segmentation.
Our methods are conceptually simple and leverage state-of-the-art instance~\cite{He2017} (or Panoptic~\cite{Por+19_cvpr,cheng2019panopticdeeplab}) segmentation models trained on \textit{existing} image datasets like Mapillary Vistas~\cite{MVD2017} for the task of object instance segmentation.
We further describe how to automatically mine tracking labels for all detected object instances, building upon state-of-the-art optical flow models~\cite{Yin_2019_CVPR}, that in turn have been trained on automatically harvested flow supervision obtained from image-based 3d modeling (structure-from-motion). 


Another important contribution of our paper is \motsnet, a deep learning, tracking-by-detection approach for MOTS.
\motsnet takes advantage of a novel mask-pooling layer, which allows it to exploit instance segmentation masks to compute a representative embedding vector for each detected object.
The tracking process can then be formulated as a series of Linear Assignment Problems (LAP), optimizing a payoff function which compares detections in the learned embedding space.

We demonstrate the superior quality for each step of our proposed MOTS training data extraction pipeline, as well as the efficacy of our novel mask-pooling layer. The key contributions and findings of our approach are:
\begin{itemize}
    \item Automated MOTS training data generation -- instance detection, segmentation and actual tracking -- is possible at scale and with very high fidelity, approximating the quality of previous state-of-the-art results~\cite{Voigtlaender_2019_CVPR} without even directly training for it.
    \item Deep-learning based, supervised optical flow~\cite{Yin_2019_CVPR} can be effectively trained on pixel-based correspondences obtained from Structure-from-Motion and furthermore used for tracklet extraction.
    \item Direct exploitation of instance segmentation masks through our proposed mask-pooling layer yields to significant improvements of sMOTSA scores
    \item The combination of our general-purpose MOTS data generation pipeline with our proposed \motsnet improves by up to 1.2\% for cars and 7.5\% for pedestrians over the previously best method on \KM in absolute terms, while using $13.7\%$ fewer parameters. 
\end{itemize}
We provide extensive ablation studies on the \KM, MOTSChallenge, and Berkeley Deep Drive BDD100k~\cite{Yu2018} datasets and obtain results consistently improving with a large margin over the prior state-of-the-art~\cite{Voigtlaender_2019_CVPR}.

\section{Related Works}
Since the MOTS task was only very recently introduced in~\cite{Voigtlaender_2019_CVPR}, directly related works are scarce.
Instead, the bulk of related works in a broader sense comes from the MOT (multi-object tracking) and VOS (video object segmentation) literature.
This is true in terms of both, datasets (see \eg~\cite{Geiger2012CVPR,MOTChallenge2015,MilanL0RS16} for MOT and \cite{Perazzi_2016_CVPR,Prest_2012_CVPR,Li_2013_ICCV} for VOS/VOT, respectively), and methods tackling the problems (\cite{sharma2018beyond} for MOT and \cite{yang2019video,Luiten_2019_ICCV,Osep18ICRA,wang2019fast} for VOS).

\paragraph{Combining motion and semantics.}
The work in~\cite{NIPS2017_7267} uses semantic segmentation results from~\cite{Long2015} for video frames together with optical flow predictions from EpicFlow~\cite{RevaudWHS15} to serve as \textit{ground truth}, for training a joint model on future scene parsing, \ie the task of anticipating future motion of scene semantics.
In~\cite{Luc_2018} this task is extended to deal with future instance segmentation prediction, based on convolutional features from a Mask R-CNN~\cite{He2017} instance segmentation branch.
In~\cite{HurR16} a method for jointly estimating optical flow and temporally consistent semantic segmentation from monocular video is introduced.
Their approach is based on a piece-wise flow model, enriched with semantic information through label consistency of superpixels.


\paragraph{Semi-automated dataset generation.}
Current MOT~\cite{wen2015ua,MilanL0RS16,Geiger2012CVPR} and VOS~\cite{xu2018youtube,Pont-Tuset_arXiv_2017} benchmarks are annotated based on human efforts.
Some of them, however, use some kind of automation.
In~\cite{xu2018youtube}, they exploit the temporal correlation between consecutive frames in a skip-frame strategy, although the segmentation masks are manually annotated.
A semi-automatic approach is proposed in~\cite{berg2019semi} for annotating multi-modal benchmarks.
It uses a pre-trained VOS model~\cite{johnander2019generative} but still requires additional manual annotations and supervision on the target dataset.
The MOTS training data from~\cite{Voigtlaender_2019_CVPR} is generated by augmenting an existing MOT dataset~\cite{MilanL0RS16,Geiger2012CVPR} with segmentation masks, using the iterative, human-in-the-loop procedure we briefly described in Sec.~\ref{sec:intro}.
The tracking labels on \cite{MilanL0RS16,Geiger2012CVPR} were manually annotated and, in the case of \cite{MilanL0RS16}, refined by adding high-confidence detections from a pedestrian detector.

The procedures used in these semi-automatic annotation approaches often resemble those of VOS systems which require user input to select objects of interest in a video.
Some proposed techniques in this field~\cite{chen2018blazingly, caelles20182018, maninis2018deep} are extensible to multiple object scenarios.
A pixel-level embedding is built in~\cite{chen2018blazingly} to solve segmentation and achieve state-of-the-art accuracy with low annotation and computational cost.

To the best of our knowledge, the only work that generates tracking data from unlabeled large-scale videos is found in~\cite{OsepVoigtlaender18ECCVW}.
By using an adaptation of the approach in~\cite{Osep18ICRA}, Osep~\etal perform object discovery on unlabeled videos from the KITTI Raw~\cite{Gei+13} and Oxford RobotCar~\cite{maddern20171} datasets.
Their approach however cannot provide joint annotations for instance segmentation and tracking.

\paragraph{Deep Learning methods for MOT and MOTS.}
Many deep learning methods for MOTS/VOS are based on ``tracking by detection'', \ie candidate objects are first detected in each frame, and then joined into tracks as a post-processing step (\eg~\cite{Voigtlaender_2019_CVPR}).
A similar architecture, \textit{i.e.} a Mask R-CNN augmented with a tracking head, is used in~\cite{yang2019video} to tackle Video Instance Segmentation.
A later work~\cite{Luiten_2019_ICCV} improved over~\cite{yang2019video} by adapting the UnOVOST~\cite{ZulfikarLuitenUnOVOST} model that, first in an unsupervised setting, uses optical flow to build short tracklets later merged by a learned embedding.
For the plain MOT task, Sharma~\etal~\cite{sharma2018beyond} incorporated shape and geometry priors into the tracking process.

Other works instead, perform detection-free tracking.
CAMOT~\cite{Osep18ICRA} tracks category-agnostic mask proposals across frames, exploiting both appearance and geometry cues from scene flow and visual odometry.
Finally, the VOS work in~\cite{wang2019fast}, follows a correlation-based approach, extending popular tracking models~\cite{bertinetto2016fully, li2018high}.
Their SiamMask model only requires bounding box initialization but is pre-trained on multiple, human-annotated datasets~\cite{LinMSCOCO2014, xu2018youtube, Rus+15}.

\section{Dataset Generation Pipeline}
\label{sec:datagen}

Our proposed data generation approach is rather generic \wrt~its data source, and here we focus on the \KR~\cite{Gei+13} dataset.
\KR contains 142 sequences (we excluded the 9 sequences overlapping with the validation set of \KM), for a total of $\sim 44$k images, captured with a professional rig including stereo cameras, LiDAR, GPS and IMU. Next we describe our pipeline which, relying only on monocular images and GPS data is able to generate accurate MOTS annotations for the dataset.

\subsection{Generation of Instance Segmentation Results}
\label{sec:instance-seg}
We begin with segmenting object instances in each frame per video and consider a predefined set $\set Y$ of $37$ object classes that belong to the Mapillary Vistas dataset~\cite{MVD2017}.
To extract the instance segments, we run the Seamless Scene Segmentation method~\cite{Por+19_cvpr}, augmented with a ResNeXt-101-32$\times$8d~\cite{Xie2016} backbone and trained on Mapillary Vistas.
By doing so, we obtain a set of object segments $\set S$ per video sequence. For each segment $s\in\set S$, we denote by $t_s\geq 0$ the frame where the segment was extracted, by $y_s\in\set Y$ the class it belongs to and by $\phi_s:\mathbb R^2\to\{0,1\}$ a pixel indicator function representing the segmentation mask, \ie $\phi_s(i,j)=1$ if and only if pixel $(i,j)$ belongs to the segment. For convenience, we also introduce the notation $\set S_t$ to denote the set of all segments extracted from frame $t$ of a given video. 
For \KR we roughly extracted 1.25M segments.

\subsection{Generation of Tracklets using Optical Flow}

\label{sec:tracklets}
After having automatically extracted the instance segments from a given video in the dataset, our goal is to leverage on optical flow to extract tracklets, \ie consecutive sequences of frames where a given object instance appears.

\paragraph{Flow training data generation.}
We automatically generate ground-truth data for training an optical flow network by running a Structure-from-Motion (SfM) pipeline, namely OpenSfM\footnote{\url{https://github.com/mapillary/OpenSfM}}, on the \KR video sequences and densify the result using PatchMatch~\cite{shen2013accurate}.
To further improve the quality of the 3d reconstruction, we exploit the semantic information that has been already extracted per frame to remove spurious correspondences generated by moving objects.
Consistency checks are also performed in order to retain correspondences that are supported by at least $3$ images.
Finally, we derive optical flow vectors between pairs of consecutive video frames in terms of the relative position of correspondences in both frames.
This process produces sparse optical flow, which will be densified in the next step.

\paragraph{Flow network.}
We train a modified version of the HD$^3$ flow network~\cite{Yin_2019_CVPR} on the dataset that we generated from \KR, without any form of pre-training.
The main differences with respect to the original implementation are the use of In-Place Activated Batch Norm (iABN)~\cite{RotPorKon18a}, which provides memory savings that enable the second difference, namely the joint training of forward and backward flow. 
We then run our trained flow network on pairs of consecutive frames in \KR in order to determine a \emph{dense} pixel-to-pixel mapping between them.
In more detail, for each frame $t$ we compute the backward mapping $\overleftarrow f_{t}:\mathbb R^2\to\mathbb R^2$, which provides for each pixel of frame $t$ the corresponding pixel in frame $t-1$.

\begin{figure*}
\centering
\includegraphics[width=\textwidth]{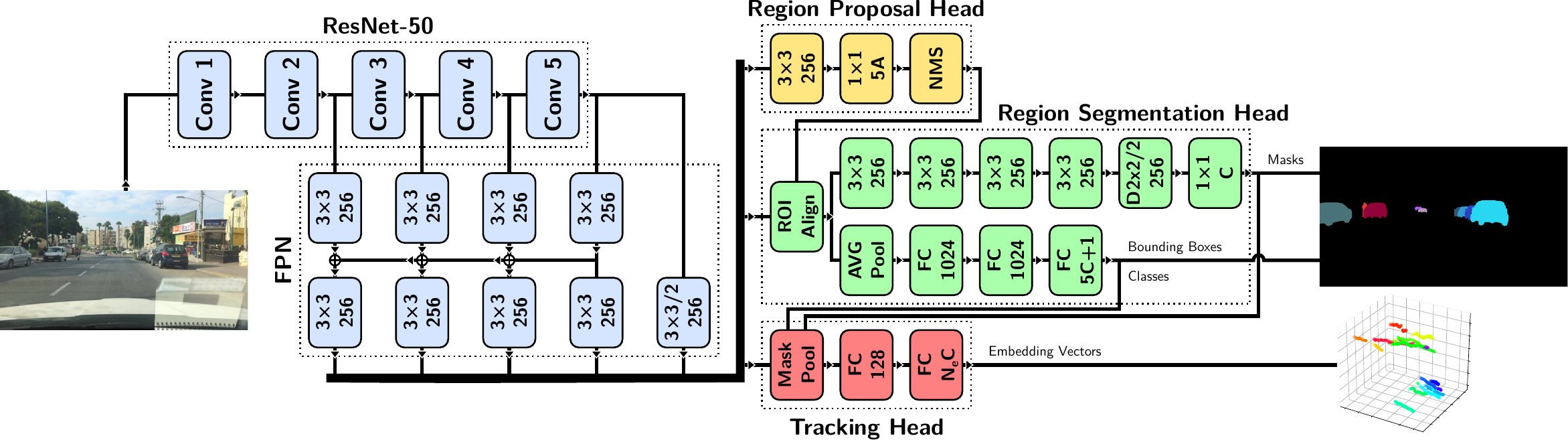}
\caption{Overview of the \motsnet architecture. Blue: network backbone; yellow: Region Proposal Head; green: Region Segmentation Head; red: Tracking Head. For an in-depth description of the various components refer to Sec.~\ref{sec:architecture} in the text.}
\label{fig:architecture}
\end{figure*}

\paragraph{Tracklet generation.} In order to represent the tracklets that we have collected up to frame $t$, we use a graph $\set G_t=(\set V_t, \set T_t)$, where vertices are all segments that have been extracted until the $t$th frame, \ie $\set V_t=\bigcup_{j\geq 1}^t\mathcal S_j$, and edges $\set T_t\subseteq \set V_t^2$ provide the matching segments across consecutive frames. We start by initializing the graph in the first frame as $\set G_1=(\set S_1, \emptyset)$. 
We construct $\set G_{t}$ for $t>1$ inductively from $\set G_{t-1}$ using the segments $\set S_t$ extracted at time $t$ and the mapping $\overleftarrow f_{t}$ according to the following procedure. The vertex set $\set V_t$ of graph $\set G_t$ is given by the union of the vertex set $\set V_{t-1}$ of $\set G_{t-1}$ and $\set S_t$, \ie $\set V_{t}=\set V_{t-1}\cup \set S_t$. The edge set is computed by solving a linear assignment problem between $\set S_{t-1}$ and $\set S_t$, where the payoff function is constructed by comparing segments in $\set S_t$ against segments in $\set S_{t-1}$ warped to frame $t$ via the mapping $\overleftarrow f_{t}$. 
We design the payoff function $\pi(\hat s, s)$ between a segment $\hat s\in\set S_{t-1}$ in frame $t-1$ and a segment $s\in\set S_t$ in frame $t$ as follows:
\[
	\pi(\hat s, s)=\text{IoU}(\phi_{s},\phi_{\hat s}\circ\overleftarrow f_{t})+\eta(\hat s, s)\,,
\]
where $\text{IoU}(\cdot,\cdot)$ computes the Intersection-over-Union of the two masks given as input, and $\eta(\hat s, s)\in\{-\infty,0\}$ is a characteristic function imposing constraints regarding valid mappings. Specifically, $\eta(\hat s, s)$ returns $0$ if and only if
the two segments belong to the same category, \ie $y_{\hat s}=y_s$, and none of the following conditions on $s$ hold:
\begin{itemize}
\item $b_1(s)-b_2(s)<\tau_0$
\item $b_1(s)<\tau_1$, and 
\item $b_1(s)/r(s)<\tau_2$, 
\end{itemize}
where $b_1(s)$ and $b_2(s)$ denote the largest and second-largest areas of intersection with $s$ obtained by segments warped from frame $t-1$ and having class $y_s$, 
and $r(s)$ denotes the area of $s$ that is not covered by segments of class $y_s$  warped from frame $t-1$.
We solve the linear assignment problem using the Hungarian algorithm, by maximizing the total payoff under a relaxed assignment constraint, which allows segments to remain unassigned. In other terms, we solve the following optimization problem:
\begin{align}
  \label{eq:optim}
	\text{maximize}&\sum_{s\in\set{S}_t}\sum_{\hat{s}\in\set{S}_{t-1}} \pi(\hat s, s) \alpha(\hat s, s)\,\\
  \label{eq:constr1}
	\text{s.t.}& \mathbox{A}{\sum_{s\in\set{S}_t}} \alpha(\hat s,s)\leq 1\qquad \forall \hat s\in\set S_{t-1}\\
  \label{eq:constr2}
	& \mathbox{A}{\sum_{\hat{s}\in\set{S}_{t-1}}} \alpha(\hat s,s)\leq 1\qquad \forall s\in\set S_{t}\\
  \label{eq:constr3}
	&\alpha(\hat s, s)\in\{0,1\}\quad\forall (\hat s, s)\in\set S_{t-1}\times\set S_t\,.
\end{align}
Finally, the solution $\alpha$ of the assignment problem can be used to update the set of edges $\set T_t$, \ie $\set T_t=\set T_{t-1}\cup \alpha^{-1}(1)$, where $\alpha^{-1}(1)$ is the set of matching segments, \ie the pre-image of $1$ under $\alpha$.

Note that this algorithm cannot track objects across full occlusions, since optical flow is unable to infer the trajectories of invisible objects.
When training \motsnet however, we easily overcome this limitation with a simple heuristic, described in Sec.~\ref{sec:training}, without needing to join the tracklets into full tracks.



\section{\motsnet}\label{sec:main}

Here we describe our multi-object tracking and segmentation approach.
Its main component is \motsnet (see Sec.~\ref{sec:architecture}), a deep net based on the Mask R-CNN~\cite{He2017} framework.
Given an RGB image, \motsnet outputs a set of instance segments, each augmented with a learned embedding vector which represents the object's identity in the sequence (see Sec.~\ref{sec:training}).
The object tracks are then reconstructed by applying a LAP-based algorithm, described in Sec.~\ref{sec:inference}.

\subsection{Architecture}
\label{sec:architecture}

\motsnet's architecture follows closely the implementation of Mask R-CNN described in~\cite{Por+19_cvpr}, but addtionally comprises a Tracking Head (TH) that runs in parallel to the Region Segmentation Head (RSH) (see Figure~\ref{fig:architecture}). 
The ``backbone'' of the network is composed of an FPN~\cite{Lin2016} component on top of a ResNet-50 body~\cite{He2015b}, producing multi-scale features at five different resolutions.
These features are fed to a Region Proposal Head (RPH), which predicts candidate bounding boxes where object instances could be located in the image.
Instance specific features are then extracted from the FPN levels using ROI Align~\cite{He2017} in the areas defined by the candidate boxes, and fed to the RSH and the TH.
Finally, for each region, the RSH predicts a vector of class probabilities, a refined bounding box and an instance segmentation mask.
Synchronized InPlace-ABN~\cite{RotPorKon18a} is employed throughout the network after each layer with learnable parameters, except for those producing output predictions.
For an in-depth description of these components we refer the reader to~\cite{Por+19_cvpr}.

\paragraph{Tracking Head (TH).}
Instance segments, together with the corresponding ROI Align features from the FPN, are fed to the TH to produce a set of $N_e$-dimensional embedding vectors.
As a first step, the TH applies ``mask-pooling'' (described in the section below) to the input spatial features, obtaining 256-dimensional vector features.
This operation is followed by a fully connected layer with 128 channels and synchronized InPlace-ABN.
Similarly to masks and bounding boxes in the RSH, embedding vectors are predicted in a class-specific manner by a final fully connected layer with $N_e\times C$ outputs, where $C$ is the number of classes, followed by L2-normalization to constrain the vectors on the unit hyper-sphere.
The output vectors are learned such that instances of the same object in a sequence are mapped close to each other in the embedding space, while instances of other objects are mapped far away.
This is achieved by minimizing a batch hard triplet loss, described in Sec.~\ref{sec:training}.

\paragraph{Mask-pooling.}
ROI Align features include both foreground and background information, but, in general, only the foreground is actually useful for recognizing an object across frames.
In our setting, by doing instance segmentation together with tracking, we have a source of information readily available to discriminate between the object and its background.
This can be exploited in a straightforward attention mechanism: pooling under the segmentation mask.
Formally, given the pixel indicator function $\phi_s(i,j)$ for a segment $s$ (see Sec.~\ref{sec:instance-seg}), and the corresponding input $N$-dimensional ROI-pooled feature map $\mathrm{X}_s:\mathbb{R}^2\to\mathbb{R}^N$, mask-pooling computes a feature vector $x_s\in\mathbb{R}^N$ as:
\[
  x_s = \frac{\sum_i\sum_j\phi_s(i, j)\mathrm{X}_s(i, j)}{\sum_i\sum_j\phi_s(i, j)}\,.
\]
During training we pool under the ground truth segmentation masks, while at inference time we switch to the masks predicted by the RSH.

\subsection{Training Losses}
\label{sec:training}

\motsnet is trained by minimizing the following loss function:
\[
  L = L_\text{TH} + \lambda (L_\text{RPH} + L_\text{RSH})\,,
\]
where $L_\text{RPH}$ and $L_\text{RSH}$ represent the Region Proposal Head and Region Segmentation Head losses as defined in~\cite{Por+19_cvpr}, and $\lambda$ is a weighting parameter.

$L_\text{TH}$ is the loss component associated with the Tracking Head, \ie the following batch hard triplet loss~\cite{hermans2017defense}:
\begin{multline*}
  L_\text{TH} = \frac{1}{|\set{S}_B|} \sum_{s\in\set{S}_B}
    \max(\max_{\hat{s} \in \mu_B(s)} \|a_s^{y_s} - a_{\hat{s}}^{y_{\hat{s}}}\| \\
    - \min_{\hat{s} \in \bar{\mu}_B(s)} \|a_s^{y_s} - a_{\hat{s}}^{y_{\hat{s}}}\| + \beta,0)\,,
\end{multline*}
where $\set{S}_B$ is the set of predicted, positive segments in the current batch, $a_s^{y_s}$ is the class-specific embedding vector predicted by the TH for a certain segment $s$ and ground truth class $y_s$, and $\beta$ is a margin parameter.
The definition of a ``positive'' segment follows the same logic as in the RSH (see~\cite{Por+19_cvpr}), \ie a segment is positive if its bounding box has high IoU with a ground truth segment's bounding box.
The functions $\mu_B(s)$ and $\bar{\mu}_B(s)$ map $s$ to the sets of its ``matching'' and ``non-matching'' segments in $\set{S}_B$, respectively.
These are defined as:
\begin{align*}
  \mu_B(s) &= \left\{\hat{s}\in\set{S}_B \mid y_s = y_{\hat{s}} \land M(s, \hat{s})\right\}\,,\\
  \bar{\mu}_B(s) &= \left\{\hat{s}\in\set{S}_B \mid y_s = y_{\hat{s}} \land \neg M(s, \hat{s})\right\}\,,
\end{align*}
where $M(s, \hat{s})$ is true iff $s$ and $\hat{s}$ belong to the same tracklet in the sequence.
Note that we are restricting the loss to only compare embedding vectors of segments belonging to the same class, effectively letting the network learn a set of class-specific embedding spaces.

In order to overcome the issue due to occlusions mentioned in Sec.~\ref{sec:tracklets}, when training on \KS we also apply the following heuristic.
Ground truth segments are only considered in the tracking loss if they are associated with a tracklet that appears in more than half of the frames in the current batch.
This ensures that, when an occlusion causes a track to split in multiple tracklets, these are never mistakenly treated by the network as different objects.

\subsection{Inference}
\label{sec:inference}

During inference, we feed frames $t$ through the network, obtaining a set of predicted segments $\set{S}_t$, each with its predicted class $y_s$ and embedding vector $a_s^{y_s}$.
Our objective is now to match the segments across frames in order to reconstruct a track for each object in the sequence.
To do this, we follow the algorithmic framework described in Sec.~\ref{sec:tracklets}, with a couple of modifications.
First, we allow matching segments between the current frame and a sliding window of $N_w$ frames in the past, in order to be able to ``jump'' across occlusions.
Second, we do not rely on optical flow anymore, instead measuring the similarity between segments in terms of their distance in the embedding space and temporal offset.

More formally, we redefine the payoff function in Eq.~\eqref{eq:optim} as:
\begin{align}
  \pi(\hat{s}, s) &= -\pi^*(\hat{s}, s) + \eta(s, \hat{s})\,,\\
  \pi^*(\hat{s}, s) &= \|a_s^{y_s} - a_{\hat{s}}^{y_{\hat{s}}}\| + \frac{|t_s - t_{\hat{s}}|}{N_w}\,,
  \label{eq:inference-po}
\end{align}
where the characteristic function $\eta(s, \hat{s})$ is redefined as:
\[
  \eta(s, \hat{s}) = \begin{cases}
    0 & \text{if}\,y_s = y_{\hat{s}} \land \pi^*(\hat{s}, s) \leq \tau\,, \\
    -\infty & \text{otherwise}\,,
  \end{cases}
\]
and $\tau$ is a configurable threshold value.
Furthermore, we replace $\set{S}_{t-1}$ in Eq.~\eqref{eq:constr1} and Eq.~\eqref{eq:constr3} with the set:
\[
  \set{U}_t = \left\{s\in \mathcal V_{t-1} \mid t_s \geq t-N_w \land (s,\hat s)\notin\set T_{t-1} \forall \hat s\right\}\,,
\]
which contains all terminal (\ie most recent) segments of the track seen in the past $N_w$ frames.
As a final step after processing all frames in the sequence, we discard the tracks with fewer than $N_t$ segments, as, empirically, very short tracks usually arise from spurious detections.

\section{Experiments}\label{sec:experiments}
We provide a broad experimental evaluation assessing i) the quality of our automatically harvested \KS dataset on the MOTS task by evaluating it against the \KM dataset, ii) demonstrate the effectiveness of \motsnet and our proposed mask-pooling layer against strong baselines, both on \KM and MOTSChallenge, iii) demonstrate the generality of our MOTS label generation process by extending the BDD100k tracking dataset with segmentation masks to become a MOTS variant thereof and iv) provide an ablation about the contribution of association terms for the final track extraction. While we can benchmark the full MOTS performance on \KM and MOTSChallenge, we are able to infer the tracking performance based on the ground truth box-based tracking annotations available for BDD100k.

\begin{table*}[th!]
    \centering
    \resizebox{0.85\textwidth}{!}{
    \begin{tabular}{l|c|cccccc|cc}
    \toprule
           & & \multicolumn{2}{c}{\textbf{sMOTSA}} & \multicolumn{2}{c}{MOTSA} & \multicolumn{2}{c|}{MOTSP} & \multicolumn{2}{c}{mAP} \\
    Method & Pre-training & Car & Ped & Car & Ped & Car & Ped & Box & Mask \\
    \midrule
    \KS (val) + HD$^3$~\cite{Yin_2019_CVPR} model zoo & inference only & 65.4 & 45.7 & 77.3 & 66.3 & \textbf{87.6} & 76.6 & -- & -- \\
    \KS (val) + HD$^3$, KITTI-SfM                     & inference only & 65.5 & 45.4 & 77.4 & 66.0 & \textbf{87.6} & 76.6 & -- & -- \\
    \midrule
    \motsnet with: & & & & & & & & \\
    \multicolumn{1}{@{\quad\quad\quad}l|}{\textsc{AveBox+TH}} &
    I & 73.7 & 46.4 & 85.8 & 62.8 & 86.7 & 76.7 & 57.4 & 50.9 \\
    \multicolumn{1}{@{\quad\quad\quad}l|}{\textsc{AveMsk-TH}} &
    I & 76.4 & 44.0 & 88.5 & 60.3 & 86.8 & 76.6 & 57.8 & 51.3 \\
    \multicolumn{1}{@{\quad\quad\quad}l|}{\textsc{AveBox-TH}} &
    I & 75.4 & 44.5 & 87.3 & 60.8 & 86.9 & 76.7 & 57.5 & 51.0 \\
    \multicolumn{1}{@{\quad\quad\quad}l|}{\KM train sequences only} &
    I & 72.6 & 45.1 & 84.9 & 62.9 & 86.1 & 75.6 & 52.5 & 47.6 \\
    \midrule
    \rowcolor{mapillarygreen}
    \motsnet & I & 77.6 & 49.1 & 89.4 & 65.6 & 87.1 & 76.4 & 58.1 & 51.8 \\
    \rowcolor{mapillarygreen}
    \motsnet  & I, M & \textbf{77.8} & \textbf{54.5} & \textbf{89.7} & \textbf{70.9} & 87.1 & \textbf{78.2} & \textbf{60.8} & \textbf{54.1} \\
    \bottomrule
    \end{tabular}}
    \caption{Results on the \KM validation set when training on \KS. First section: ablation results. Second section: main results and comparison with state of the art. Note that all \motsnet variants are trained exclusively on machine generated annotations, while TrackR-CNN is trained on human annotated ground truth.}
    \label{tab:results-ks}
    \end{table*}

\subsection{Evaluation Measures}
The CLEAR MOT~\cite{bernardin2008evaluating} metrics, including Multi-Object Tracking Accuracy (MOTA) and Precision (MOTP), are well established as a standard set of measures for evaluating Multi-Object Tracking systems.
These, however, only account for the bounding box of the tracked objects.
Voigtlaender~\etal~\cite{Voigtlaender_2019_CVPR} describe an extension of MOTA and MOTP that measure segmentation as well as tracking accuracy, proposing the sMOTSA, MOTSA and MOTSP metrics.
In particular, MOTSA and MOTSP are direct equivalents of MOTA and MOTP, respectively, where the prediction-to-ground-truth matching process is formulated in terms of mask IoU instead of bounding box IoU.
Finally, sMOTSA can be regarded as a ``soft'' version of MOTSA where the contribution of each true positive segment is weighted by its mask IoU with the corresponding ground truth segment.
Please refer to Appendix~\ref{sec:metrics} for a formal description of these metrics.

\subsection{Experimental Setup}
\label{sec:exp-setup}
All the results in the following sections are based on the same \motsnet configuration, with a ResNet-50 backbone and embedding dimensionality fixed to $N_e=32$.
Batches are formed by sampling a subset of $N_w$ full-resolution, contiguous frames at a random offset in one of the training sequences, for each GPU.
During training we apply scale augmentation in the $[0.8,1.25]$ range and random flipping.
Training by SGD follows the following linear schedule: $\text{lr}_i=\text{lr}_0(1-\frac{i}{\#\text{steps}})$, where $\text{lr}_0$ is the starting learning rate and $i$ is the training step (i.e. batch) index.\footnote{Please refer to Appendix~\ref{sec:params} for a full specification of the training parameters.}
The network weights are initialized from an ImageNet-pretrained model (``I'' in the tables), or from a Panoptic Segmentation model trained on the Mapillary Vistas dataset (``M'' in the tables), depending on the experiment.
Differently from~\cite{Voigtlaender_2019_CVPR}, we do not pre-train on MS COCO~\cite{LinMSCOCO2014} (``C'' in the tables).

The hyper-parameters of the inference algorithm are set as follows: threshold $\tau=1$; window size $N_w$ equal to the per-GPU batch size used during training; minimum track length $N_t=5$.
Note that, in contrast with~\cite{Voigtlaender_2019_CVPR}, we do not fine-tune these parameters, instead keeping them fixed in all experiments, and independent of class.
All experiments are run on four V100 GPUs with 32GB of memory.

\subsection{\KM}
\label{sec:experiments-ks}
The \KM dataset was introduced in~\cite{Voigtlaender_2019_CVPR}, adding instance segmentation masks for cars and pedestrians to a subset of 21 sequences from KITTI raw~\cite{Geiger2012CVPR}. It has a total of 8.008 images (5.027 for training and 2.981 for validation), containing approximately 18.8k/8.1k annotated masks for 431/99 tracks of cars/pedestrians in training, and roughly 8.1k/3.3k masks for 151/68 tracks on the validation split. Approximately $12.3\%$ of all masks were manually annotated and the rest are human-verified and -corrected instance masks. As the size of this dataset is considerably larger and presumably more informative than MOTSChallenge, we focused our ablations on it. 

Table~\ref{tab:results-ks} summarizes the analyses and experiments we ran on our newly generated \KS dataset, by evaluating on the official \KM validation set. We discuss three types of ablations in what follows.

\paragraph{\KS data quality analysis.}
The two topmost rows of Tab.~\ref{tab:results-ks} show the results when generating the validation set results solely with our proposed dataset generation pipeline, described in Section~\ref{sec:datagen}, \ie no learned tracking component is involved. 
Our synthetically extracted validation data considerably outperforms the sMOTSA scores on pedestrians and almost obtains the performance on cars for CAMOT~\cite{Osep18ICRA}, as reported in~\cite{Voigtlaender_2019_CVPR}.
When training the flow network on the training data generated from SfM as described in Section~\ref{sec:datagen}, we match the performance obtained by the best-performing HD$^3$~\cite{Yin_2019_CVPR} model for KITTI\footnote{Pre-trained models available at \url{https://github.com/ucbdrive/hd3}.}.
While this validates our tracklet generation process and the way we harvest optical flow training data, we investigate further the effects of learning from imperfect data (see Sec.~\ref{sec:tracklets}).

\paragraph{\motsnet tracking head ablations.}
The center block in Tab~\ref{tab:results-ks} compares different configurations for the Tracking Head described in~\ref{sec:main}.
The first variant (\textsc{AveBox+TH}) replaces the mask-pooling operation in the Tracking Head with average pooling on the whole box.
The second and third variants (\textsc{AveMsk-TH}, \textsc{AveBox-TH}) completely remove the Tracking Head and its associated loss, instead directly computing the embeddings by pooling features under the detected mask or box, respectively.
All variants perform reasonably well and improve over~\cite{Osep18ICRA,osep2017combined} on the primary sMOTSA metric.
Notably, \textsc{AveMsk-TH}, \ie the variant using our mask-pooling layer and no tracking head, is about on par with TrackR-CNN on cars despite being pre-trained only on ImageNet, using a smaller, ResNet-50 backbone and not exploiting any tracking supervision.
The last variant in this block shows the performance obtained when \motsnet is trained only on images also in the \KM training set, and with our full model.
Interestingly, the scores only drop to an extent where the gap to~\cite{Voigtlaender_2019_CVPR} might be attributed to their additional pre-training on MS COCO and Mapillary Vistas and a larger backbone (\textit{cf.}~Tab.~\ref{tab:results-km}). Finally, comparing the latter directly to our \motsnet results from Tab.~\ref{tab:results-km} (ImageNet pre-trained only), where we directly trained on \KM training data, our \KS{}-based results improve substantially on cars, again confirming the quality of our automatically extracted dataset.

The bottommost part of this table shows the performance when training on the full \KS dataset and evaluating on \KM validation, using different pre-training settings.
The first is pre-trained on ImageNet and the second on both, ImageNet and Mapillary Vistas.
While there is only little gap between them on the car class, the performance on pedestrians rises by over 5\% when pre-training on Vistas.
The most encouraging finding is that using our solely automatically extracted \KS dataset we obtain significantly improved scores (+1.6\% on cars, +7.4\% on pedestrians) in comparison to the previous state-of-the-art method from~\cite{Voigtlaender_2019_CVPR}, trained on manually curated data.

\paragraph{\motsnet on \KM.}
In Tab.~\ref{tab:results-km} we directly compare against previously top-performing works including TrackR-CNN~\cite{Voigtlaender_2019_CVPR} and references from therein, \eg CAMOT~\cite{Osep18ICRA}, CIWT~\cite{osep2017combined}, and BeyondPixels~\cite{sharma2018beyond}.
It is worth noting that the last three approaches were reported in~\cite{Voigtlaender_2019_CVPR} and partially built on.
We provide all relevant MOTS measures together with separate average AP numbers to estimate the quality of bounding box detections or instance segmentation approaches, respectively.
For our \motsnet we again show results under different pre-training settings (ImageNet, Mapillary Vistas and \KS).
Different pre-trainings affect the classes cars and pedestrians in a different way, \ie the ImageNet and Vistas pre-trained model performs better on pedestrians while the \KS pre-trained one is doing better on cars.
This is most likely due to the imbalanced distribution of samples in \KS while our variant with pre-training on both, Vistas \textit{and} \KS yields the overall best results.
We obtain absolute improvements of \textbf{1.9\%/7.5\%} for cars/pedestrians over the previously best work in~\cite{Voigtlaender_2019_CVPR} despite using a smaller backbone and no pre-training on MS COCO.
Finally, also the recognition metrics for both, detection (box mAP) and instance segmentation (mask mAP) significantly benefit from pre-training on our \KS dataset, and in conjuction with Vistas rise by 7.2\% and 6.4\% over the ImageNet pre-trained variant, respectively.

\begin{table*}
\centering
\resizebox{0.7\textwidth}{!}{
\begin{tabular}{l|c|cccccc|cc}
\toprule
& & \multicolumn{2}{c}{\textbf{sMOTSA}} & \multicolumn{2}{c}{MOTSA} & \multicolumn{2}{c|}{MOTSP} & \multicolumn{2}{c}{mAP} \\
Method & Pre-training & Car & Ped & Car & Ped & Car & Ped & Box & Mask \\
\midrule
TrackR-CNN~\cite{Voigtlaender_2019_CVPR} &
I, C, M & 76.2 & 47.1 & 87.8 & 65.5 & 87.2 & 75.7 & -- & -- \\
CAMOT~\cite{Osep18ICRA} &
I, C, M & 67.4 & 39.5 & 78.6 & 57.6 & 86.5 & 73.1 & -- & -- \\
CIWT~\cite{osep2017combined} &
I, C, M & 68.1 & 42.9 & 79.4 & 61.0 & 86.7 & 75.7 & -- & -- \\
BeyondPixels~\cite{sharma2018beyond} &
I, C, M & 76.9 & -- & \textbf{89.7} & -- & 86.5 & -- & -- & -- \\
\midrule
\multirow{4}{*}{\motsnet} & I & 69.0 & 45.4 & 78.7 & 61.8 & 88.0 & 76.5 & 55.2 & 49.3 \\
& I, M & 74.9 & 53.1 & 83.9 & 67.8 & 89.4 & 79.4 & 60.8 & 54.9 \\
& I, KS & 76.4 & 48.1 & 86.2 & 64.3 & 88.7 & 77.2 & 59.7 & 53.3 \\
\rowcolor{mapillarygreen}
& I, M, KS & \textbf{78.1} & \textbf{54.6} & 87.2 & \textbf{69.3} & \textbf{89.6} & \textbf{79.7} & \textbf{62.4} & \textbf{55.7} \\
\bottomrule
\end{tabular}}
\caption{Results on the \KM validation set when training on the \KM training set. First section: state of the art results using masks and detections from~\cite{Voigtlaender_2019_CVPR}. Second section: our results under different pre-training settings.}
\label{tab:results-km}
\end{table*}

\subsection{\motsnet on MOTSChallenge}
\label{sec:mots-challenge}

\begin{figure}
\centering
\includegraphics[width=0.32\columnwidth]{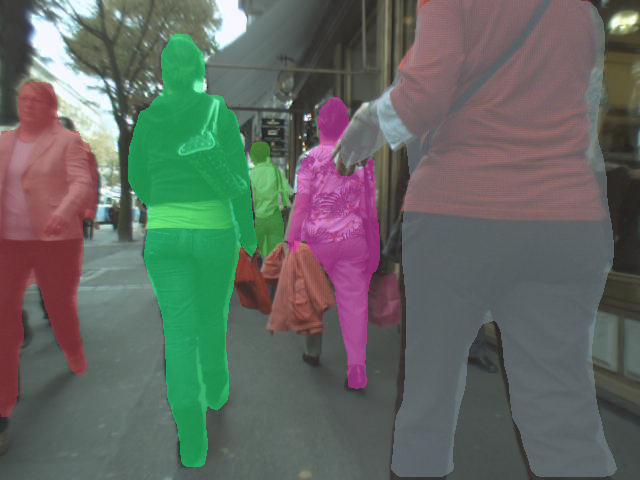}
\includegraphics[width=0.32\columnwidth]{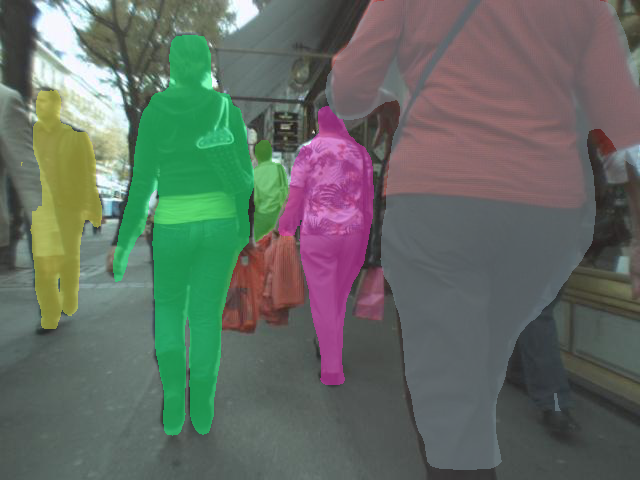}
\includegraphics[width=0.32\columnwidth]{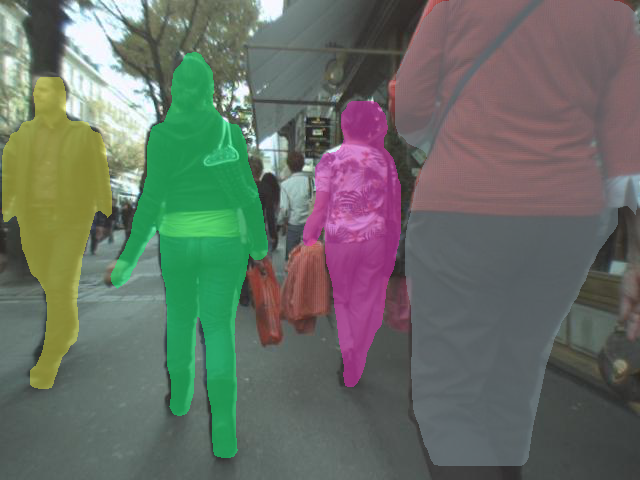}
\caption{Sample \motsnet predictions on a sub-sequence from the MOTS Challenge dataset.}
\label{fig:mots-challenge}
\end{figure}

In Tab.~\ref{tab:results-mc} we present our results on MOTSChallenge, the second dataset contributed in~\cite{Voigtlaender_2019_CVPR} and again compare against all related works reported therein.
This dataset comprises of 4 sequences, a total of 2.862 frames and 228 tracks with roughly 27k pedestrians, and is thus significantly smaller than \KM.
Due to the smaller size, the evaluation in~\cite{Voigtlaender_2019_CVPR} runs leave-one-out cross validation on a per-sequence basis.
We again report numbers for differently pre-trained versions of \motsnet.
The importance of segmentation pre-training on such small datasets is quite evident: while \motsnet (I) shows the overall worst performance, its COCO pre-trained version significantly improves over all baselines. We conjecture that this is also due to the type of scenes -- many sequences are recorded with a static camera and crossing pedestrians are shown in a quite close-up setting (see \eg Fig.~\ref{fig:mots-challenge}).

\begin{table}
\centering
\resizebox{\columnwidth}{!}{
\begin{tabular}{l|c|ccc}
\toprule
Method & Pre-training & sMOTSA & MOTSA & MOTSP \\
\midrule
TrackR-CNN~\cite{Voigtlaender_2019_CVPR} & I, C, M & 52.7 & 66.9 & 80.2 \\
MHT-DAM~\cite{kim2015multiple} & I, C, M & 48.0 & 62.7 & 79.8 \\
FWT~\cite{henschel2018fusion} & I, C, M & 49.3 & 64.0 & 79.7 \\
MOTDT~\cite{long2018real} & I, C, M & 47.8 & 61.1 & 80.0 \\
jCC~\cite{keuper2018motion} & I, C, M & 48.3 & 63.0 & 79.9 \\
\midrule
\multirow{2}{*}{\motsnet} & I & 41.8 & 55.2 & 78.4 \\
& I, C & \textbf{56.8} & \textbf{69.4} & \textbf{82.7} \\
\bottomrule
\end{tabular}}
\caption{Results on the MOTSChallenge dataset. Top section: state-of-the-art results using masks from~\cite{Voigtlaender_2019_CVPR}. Bottom section: our \motsnet results under different pre-training settings.}
\label{tab:results-mc}
\end{table}

\subsection{\motsnet on BDD100k} 
Since our MOTS training data generation pipeline can be directly ported to other datasets we also conducted experiments on the recently released BDD100k tracking dataset~\cite{Yu2018}.
It comprises 61 sequences, for a total of $\sim 11$k frames and comes with bounding box based tracking annotations for a total of 8 classes (in our evaluation we focus only on cars and pedestrians, for compatibility with \KS and \KM).
We split the available data into training and validation sets of, respectively, 50 and 11 sequences, and generate segmentation masks for each annotated bounding box, again by using~\cite{Por+19_cvpr} augmented with a ResNeXt-101-32$\times$8d backbone.
During data generation, the instance segmentation pipeline detected many object instances missing in the annotations, which is why we decided to provide \textit{ignore} box annotations for detections with very high confidence but missing in the ground truth.
Since there are only bounding-box based tracking annotations available, we present MOT rather than MOTS tracking scores.
The results are listed in Tab.~\ref{tab:results-bdd100k}, comparing pooling strategies \textsc{AveBox+Loss} and \textsc{AveMsk+Loss}, again as a function of different pre-training settings.
We again find that our proposed mask-pooling results compare favourable against the vanilla box-based pooling.
While the improvement is small compared to the ImageNet-only pre-trained backbone, the Vistas pre-trained model benefits noticeably.

\begin{table}
\centering
\resizebox{0.85\columnwidth}{!}{
\begin{tabular}{l|c|cc}
\toprule
\motsnet variant & Pre-training & \textbf{MOTA} & MOTP \\
\midrule
\multirow{2}{*}{\textsc{AveBox+Loss}} & I & 53.8 & 83.1 \\
& I, M & 56.9 & 83.9 \\
\multirow{2}{*}{\textsc{AveMsk+Loss}} & I & 53.9 & 83.1 \\
& I, M & \textbf{58.2} & \textbf{84.0} \\
\bottomrule
\end{tabular}}
\caption{MOT ablation results on the BDD100K dataset.}
\label{tab:results-bdd100k}
\end{table}

\subsection{Ablations on Track Extraction}
\label{sec:track_ablations}
\begin{figure}
\includegraphics[width=\columnwidth]{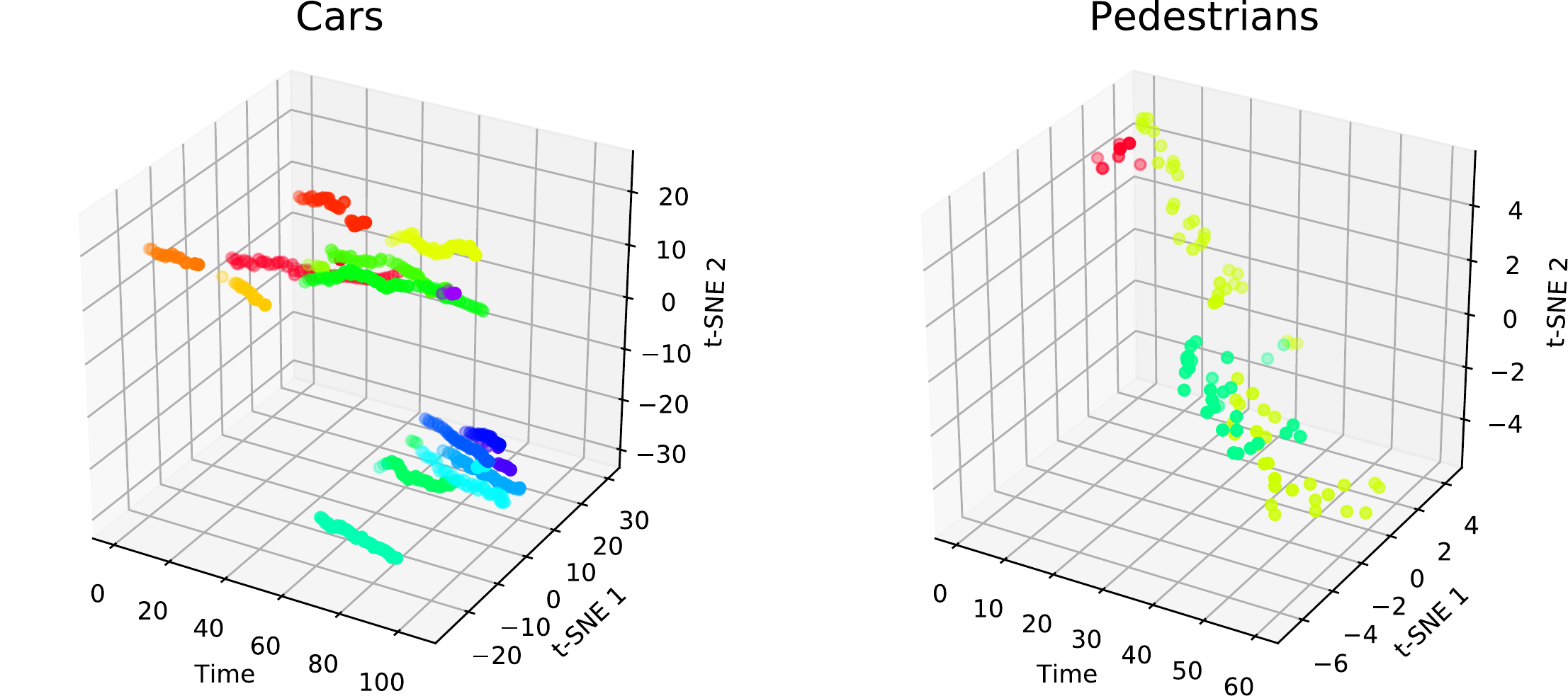}
\caption{t-SNE visualization of the embedding vectors computed by the Tracking Head for sequence ``0014'' of \KM. Points corresponding to detections of the same object have the same color.}
\label{fig:tSNE}
\end{figure}

Here we analyze the importance of different cues in the payoff function used during inference (see Section~\ref{sec:inference}): distance in the embedding space (Embedding), distance in time (Time) and signed intersection over union~\cite{simonelli2019disentangling} (sIoU) between bounding boxes\footnote{Please refer to the Appendix~\ref{sec:sIoU} for more details on the use of sIoU.}.
As a basis, we take the best performing \motsnet model from our experiments on \KM, listed at the bottom of Table~\ref{tab:results-km}.
This result was obtained by combining Embedding and Time, as in Eq.~\eqref{eq:inference-po}.
As can be seen from differently configured results in Tab.~\ref{tab:results-TrackletGraph}, the embedding itself already serves as a good cue, and can be slightly improved on pedestrians when combined with information about proximity in time (with $0.1$ drop on cars), while outperforming sIoU.
Figure~\ref{fig:tSNE} shows a visualization of the embedding vectors learned by this model.

\begin{table}
    \centering
    \resizebox{\columnwidth}{!}{
    \begin{tabular}{ccc|cc}
    \toprule
    sIoU & Embedding & Time & sMOTSA Car & sMOTSA Ped \\
    \midrule
    \cmark & \xmark & \xmark & 76.7 & 50.8 \\
    \xmark & \cmark & \xmark & 78.2 & 54.4 \\
    \cmark & \xmark & \cmark & 77.0 & 51.8 \\
    \xmark & \cmark & \cmark & 78.1 & 54.6 \\
    \bottomrule
    \end{tabular}}
    \caption{Ablation results on the \KM dataset, using our best performing model from Tab.~\ref{tab:results-km} when switching between different cues in the payoff function for inference.}
    \label{tab:results-TrackletGraph}
\end{table}

\section{Conclusions}
In this work we addressed and provided two major contributions for the novel task of multi-object tracking and segmentation (MOTS). First, we introduced an automated pipeline for extracting high-quality training data from generic street-level videos to overcome the lack of MOTS training data, without time- and cost-intensive, manual annotation efforts. Data is generated by solving the linear assignment on a causal tracking graph, where instance segmentations per frame define the nodes, and optical-flow based compatibilities represent edges as connections over time. Our second major contribution is a deep-learning based \motsnet architecture to be trained on MOTS data, exploiting a novel mask-pooling layer that guides the association process for detections based on instance segmentation masks. We provide exhaustive ablations for both, our novel training data generation process and our proposed \motsnet. We improve over the previously best work~\cite{Voigtlaender_2019_CVPR} for the \KM dataset (1.9\%/7.5\% for cars/pedestrians) and on MOTSChallenge (4.1\% for pedestrians), respectively.

\appendix

\section{CLEAR MOT and MOTS metrics}
\label{sec:metrics}

The CLEAR MOT metrics (including MOTA and MOTP) are first defined in~\cite{bernardin2008evaluating} to evaluate Multi-Object Tracking systems.
In order to compute MOTA and MOTP, a matching between the ground truth and predicted tracklets needs to be computed at each frame by solving a linear assignment problem.
We will not repeat the details of this process, instead focusing on the way its outputs are used to compute the metrics.
In particular, for each frame $t$, the matching process gives:
\begin{itemize}
  \item the number of correctly matched boxes $\text{TP}_t$;
  \item the number of mismatched boxes $\text{IDS}_t$, \ie the boxes belonging to a predicted tracklet that was matched to a different ground truth tracklet in the previous frame;
  \item the number of false positive boxes $\text{FP}_t$, \ie the predicted boxes that are not matched to any ground truth;
  \item the number of ground truth boxes $\text{GT}_t$;
  \item the intersection over union $\text{IoU}_{t,i}$ between each correctly predicted box and its matching ground truth.
\end{itemize}
Given these, the metrics are defined as:
\[
  \text{MOTA} = \frac{\sum_t (\text{TP}_t - \text{FP}_t - \text{IDS}_t)}{\sum_t \text{GT}_t}\,,
\]
and
\[
  \text{MOTP} = \frac{\sum_{t,i} \text{IoU}_{t,i}}{\sum_t \text{TP}_t}\,.
\]

The MOTS metrics~\cite{Voigtlaender_2019_CVPR} extend the CLEAR MOT metrics to the segmentation case.
Their computation follows the overall procedure described above, with a couple of exceptions.
First, box IoU is replaced with mask IoU.
Second, the matching process is simplified by defining a ground truth and predicted segment to be matching if and only if their IoU is greater than 0.5.
Different from the bounding box case, the segmentation masks are assumed to be non-overlapping, meaning that this criterion results in a unique matching without the need to solve a LAP.
With these changes, and given the definitions above, the MOTS metrics are:
\[
  \text{MOTSA} = \frac{\sum_t (\text{TP}_t - \text{FP}_t - \text{IDS}_t)}{\sum_t \text{GT}_t}\,,
\]
\[
  \text{sMOTSA} = \frac{\sum_t (\sum_i \text{IoU}_{t,i} - \text{FP}_t - \text{IDS}_t)}{\sum_t \text{GT}_t}\,,
\]
and
\[
  \text{MOTSP} = \frac{\sum_{t,i} \text{IoU}_{t,i}}{\sum_t \text{TP}_t}\,.
\]

\section{Hyper-parameters}
\label{sec:params}

\subsection{Data Generation}
In the data generation process, when constructing tracklets (see Sec.~\ref{sec:tracklets}) we use the following parameters: $\tau_0=10\text{pix}$, $\tau_1=10\text{pix}$, $\tau_2=2$.

\subsection{Network training}
As mentioned in Sec.~\ref{sec:exp-setup}, all our trainings follow a linear learning rate schedule:
\[
  \text{lr}_i = \text{lr}_0 \left(1 - \frac{i}{\#\text{steps}}\right)\,,
\]
where the initial learning rate $\text{lr}_0$ and total number of steps depend on the dataset and pre-training setting.
The actual values, together with the per-GPU batch sizes $N_w$ are reported in Tab.~\ref{tab:hyper}.
The loss weight parameter $\lambda$ in the first equation of Sec.~\ref{sec:training} is fixed to $1$ in all experiments, except for the COCO pre-trained experiment on MOTSChallenge, where $\lambda=0.1$.

\begin{table}
\resizebox{\columnwidth}{!}{
\begin{tabular}{l|c|ccc}
  \toprule
    Dataset & Pre-training & $\text{lr}_0$ & \# epochs & $N_w$ \\
  \midrule
    \multirow{2}{*}{\KS} & I & 0.02 & 20 & 12 \\
    & M & 0.01 & 10 & 12 \\
    \KS, \KM sequences & I & 0.02 & 180 & 12 \\
  \midrule
    \multirow{2}{*}{\KM} & I & 0.02 & 180 & 12 \\
    & M, KS & 0.01 & 90 & 12 \\
  \midrule
    BDD100k & all & 0.02 & 100 & 10 \\
  \midrule
    \multirow{2}{*}{MOTSChallenge} & I & 0.01 & 90 & 10 \\
    & C & 0.02 & 180 & 10 \\
  \bottomrule
\end{tabular}}
  \caption{\motsnet training hyperparameters for different datasets and pre-training settings.}
  \label{tab:hyper}
\end{table}

\section{Signed Intersection over Union}
\label{sec:sIoU}

Signed Intersection over Union, as defined in~\cite{simonelli2019disentangling}, extends standard intersection over union between bounding boxes, by providing meaningful values when the input boxes are not intersecting.
Given two bounding boxes $\hat{\vct{b}}=(\hat{u}_1, \hat{v}_1, \hat{u}_2, \hat{v}_2)$ and $\vct{b}=(u_1, v_1, u_2, v_2)$, where $(u_1, v_1)$ and $(u_2, v_2)$ are the coordinates of a box's top-left and bottom-right corners, respectively, the signed intersection over union $\text{sIoU}(\hat{\vct{b}}, \vct{b})$ is:
\begin{itemize}
  \item greater than 0 and equal to standard intersection over union when the boxes overlap;
  \item less than 0 when the boxes don't overlap, and monotonically decreasing as their distance increases.
\end{itemize}
This is obtained by defining:
\[
  \text{sIoU}(\hat{\vct{b}}, \vct{b}) = \frac{|\hat{\vct{b}}\sqcap\vct{b}|_\pm}{|\hat{\vct{b}}|+|\vct{b}|-|\hat{\vct{b}}\sqcap\vct{b}|_\pm}\,,
\]
where
\[
  \hat{\vct{b}}\sqcap\vct{b} = \left(\begin{array}{l}
    \max(\hat{u}_1, u_1) \\
    \max(\hat{v}_1, v_1) \\
    \min(\hat{u}_2, u_2) \\
    \min(\hat{v}_2, v_2)
  \end{array}\right)
\]
is an extended intersection operator, $|\vct{b}|$ denotes the area of $\vct{b}$, and
\[
  |\vct{b}|_\pm = \begin{cases}
    +|\vct{b}| & \text{if}\ u_2 > u_1 \land v_2 > v_1, \\
    -|\vct{b}| & \text{otherwise},
  \end{cases}
\]
is the ``signed area'' of $\vct{b}$.

Signed intersection over union is used in the ablation experiments of Sec.~\ref{sec:track_ablations} as an additional term in the payoff function $\pi(\hat{s}, s)$ as follows:
\begin{align*}
  \pi(\hat{s}, s) &= -\pi^*(\hat{s}, s) + \eta(s, \hat{s})\,,\\
  \pi^*(\hat{s}, s) &= \text{sIoU}(\vct{b}_s, \vct{b}_{\hat{s}}) + \|a_s^{y_s} - a_{\hat{s}}^{y_{\hat{s}}}\| + \frac{|t_s - t_{\hat{s}}|}{N_w}\,,
  \label{eq:inference-po}
\end{align*}
where $\vct{b}_s$ denotes the bounding box of segment $s$.\\

\paragraph{Acknowledgements.} Idoia and Joan acknowledge financial support by the Spanish grant TIN2017-88709-R (MINECO/AEI/FEDER, UE).

{\small
\bibliographystyle{ieee_fullname}
\bibliography{marmotsbib}{}
}

\end{document}